\documentclass[conference,letterpaper]{IEEEtran}
\usepackage[OT1]{fontenc}
\IEEEoverridecommandlockouts

\usepackage{cite}
\usepackage{graphicx}
\usepackage{tikz}
\usetikzlibrary{arrows.meta}
\usepackage{amsmath,amssymb}
\usepackage{booktabs}
\usepackage{multirow}
\usepackage{xcolor}
\usepackage{url}
\usepackage[hidelinks]{hyperref}
\usepackage{subcaption}
\usepackage{placeins}

\definecolor{safetynet}{RGB}{214,120,20}
\input{bib/bib_short.def}

\newcommand{\ppMissTotal}{80}
\newcommand{\ppMissRows}{76}

\newcommand{\ppMissFacing}{4}

\newcommand{\ours}{\textsc{FM-Anonymize}}

\title{Training-Free Anonymization of Driving Data with Promptable
Segmentation and 3D Human Mesh Recovery}

\author{
   \IEEEauthorblockN{
       Max A. Buettner, 
       Mark Azer, 
       Fabian B. Flohr
   }
   \IEEEauthorblockA{Munich University of Applied Sciences, Intelligent Vehicles Lab (IVLab), Munich, Germany\\
   \{max.buettner, mark.azer, fabian.flohr\}@hm.edu}
}

\begin{document}
\bstctlcite{IEEEexample:BSTcontrol}
\maketitle

\begin{abstract}
Camera recordings for autonomous driving contain faces and license plates
that should be anonymized before labeling, sharing, or training. Dedicated
detectors can miss small plates and faces behind vehicle glass. We present
\ours{}, a training-free anonymization pipeline that combines promptable
segmentation, 3D human mesh recovery, and generative replacement. SAM~3 locates people, faces, and
vehicles and searches zoomed vehicle crops for plates, occupants, and
windows. SAM~3D Body assesses face visibility from head orientation and projects head regions for blurring. Optional diffusion
inpainting replaces eligible blurred heads with synthetic faces,
guided by estimated pose and keypoints.
On PP4AV, the blur pipeline
protects 97.8\,\% of faces and 95.5\,\% of plates, the highest
privacy recall of all compared methods at their default settings.
On our recordings, where the localization thresholds were set, it protects 97.1\,\% of
faces and 93.4\,\% of plates, compared with 93.7\,\%/73.0\,\% for
UrbanAnonymizer, 42.7\,\%/40.7\,\% for EgoBlur, and 96.6\,\% of
faces for deface. Synthetic replacement raises full-body
keypoint average precision over blurring from 54.1\% to 79.6\% on our recordings
and from 39.9\% to 68.3\% on PP4AV. Code and configurations are
available on \href{https://github.com/max-a-ai/FM-Anonymize}{\textcolor{blue}{github}}.
\end{abstract}

\section{Introduction}
\begin{figure}[!t]
\centering
\input{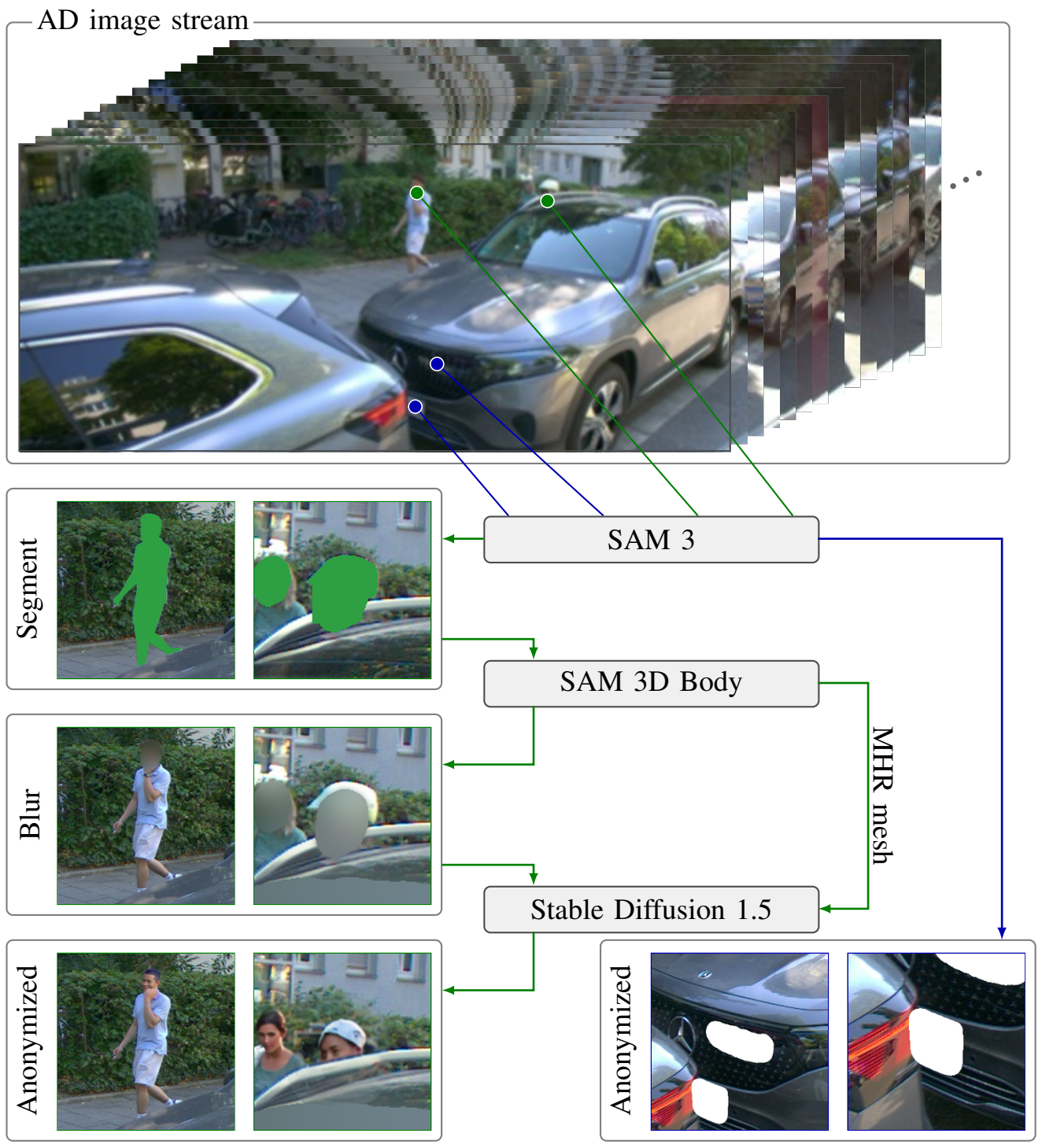}
\caption{\ours{} at a glance on a frame of an AD image stream. SAM~3
segments persons (green) and plates (blue). SAM~3D Body finds each
person's head, whose region is blurred. Optionally, Stable Diffusion~1.5, guided
by the MHR mesh, paints a new face into each eligible head. Plates, here a reflected and
a partial one, are filled white.}

\label{fig:header}
\end{figure}
Autonomous driving (AD) relies on recorded sensor data that often
captures identifiable people: pedestrians and cyclists, drivers
behind windshields, and individuals linked to vehicles through
license plates. Such recordings raise data-protection concerns
under the GDPR~\cite{gdpr2016,edpb2020video}, while the EU AI Act
places data-governance requirements on high-risk AI
systems~\cite{aiact2024}. We therefore target anonymization before
recordings enter annotation, sharing, or training workflows,
reducing exposure of personal data from the outset. Protecting
privacy, however, must be balanced with preserving the information
that makes these recordings useful. Blurring faces can obscure
head orientation and facial keypoints needed by labelers and
perception models~\cite{hukkelas2023doesanon}. The challenge is thus
to suppress identifying information while retaining the visual
cues needed to interpret road users and their behavior.

Existing anonymization pipelines differ in how they locate and
modify sensitive regions. Public driving datasets use dedicated
processing pipelines: nuScenes filters face and plate detections
using projected 3D boxes~\cite{caesar2020nuscenes}, A2D2 uses a
segmentation network to blur plates and pedestrian
heads~\cite{geyer2020a2d2}, and Argoverse~2 blurs faces and
plates~\cite{wilson2021argoverse2}. Open-source tools such as
deface~\cite{deface}, EgoBlur~\cite{raina2023egoblur}, and
UrbanAnonymizer~\cite{turkcan2026urbananonymizer} rely on dedicated
detectors. These approaches can miss small or partly occluded
targets, while enlarging
the anonymized regions can remove useful scene
content~\cite{hukkelas2023doesanon}. Generative methods instead
replace sensitive regions with synthetic content to retain a more
natural appearance~\cite{hukkelas2019deepprivacy,hukkelas2023deepprivacy2,klemp2023ldfa,wang2026diffanon},
but still depend on reliable localization.
This leaves two connected challenges: finding regions that require
protection and modifying them while preserving information useful
for perception.

We address these challenges by \emph{instructing} pretrained
foundation models rather than training task-specific detectors, through \ours{}. 
A text-prompted segmentation model,
SAM~3~\cite{carion2026sam3}, locates people and vehicles and
searches zoomed vehicle crops for license plates, occupants, and
reflections. SAM~3D Body~\cite{yang2026sam3dbody} recovers human meshes
using the camera intrinsics. The estimated head
orientation determines whether a face is likely visible, and the
projected head defines the region to anonymize. Plates are filled
white, while faces and potentially identifying reflective regions are
blurred. To retain useful visual cues, an optional generative step
replaces eligible blurred head and occupant regions with synthetic
faces. Diffusion inpainting uses the estimated head pose and
keypoints to guide replacement, linking region
localization and content generation through the same geometric
estimates. Our contributions are:
\begin{itemize}
\item a training-free anonymization pipeline for
driving data that combines promptable segmentation, zoomed vehicle
searches, and human mesh recovery to localize sensitive regions
(Sec.~\ref{sec:method}).
\item a synthetic face replacement step that reuses
the estimated head geometry to guide diffusion inpainting,
aiming to preserve pose and keypoint information (Sec.~\ref{sec:method}-F).
\item an evaluation on PP4AV~\cite{trinh2023pp4av} and on our own recordings from a car and a micro-mobility agent, comparing
privacy coverage against open-source baselines and assessing
re-identification, visual realism, and pose preservation for
blurring and synthetic replacement, with ablations of the
localization pipeline (Sec.~\ref{sec:exp}).
\end{itemize}

\section{Related Work}
\label{sec:related}
\textbf{Anonymization tools and practice.}
Google Street View first blurred street imagery at scale with face and
plate detectors~\cite{frome2009streetview}. Driving
datasets followed with their own pipelines:
nuScenes~\cite{caesar2020nuscenes} runs plate and face detectors at a
high-recall threshold and keeps only detections that overlap projected
3D boxes, A2D2~\cite{geyer2020a2d2} blurs plates and pedestrian heads
with a semantic segmentation network, and
Argoverse~2~\cite{wilson2021argoverse2} blurs faces and plates.
Our method shares the idea of reasoning inside a parent object with
nuScenes and A2D2, but replaces box overlap by promptable segmentation and a 3D
head model, so it knows \emph{whether} a face is visible, and it needs
no labeled training data. Openly available tools are deface~\cite{deface}
(CenterFace~\cite{xu2019centerface}, faces only),
EgoBlur~\cite{raina2023egoblur} (Faster R-CNN detectors for faces and
plates, developed for egocentric Aria recordings), and the archived
understand.ai anonymizer~\cite{uaianonymizer}.
PP4AV~\cite{trinh2023pp4av}
provides face and plate ground truth for driving scenes. Compared to these, our
approach needs no training data, runs on premises, is open source for
research, and is set up by writing text prompts instead of training
detectors.
\textbf{Generative anonymization for driving data.}
Generative replacement methods such as
DeepPrivacy2~\cite{hukkelas2023deepprivacy2} and
LDFA~\cite{klemp2023ldfa} preserve a more natural image appearance.
DeepPrivacy2 combines DSFD face detection with CSE-based full-body
anonymization, whereas LDFA uses RetinaFace followed by pretrained
diffusion inpainting for driving scenes. Their face-generation
stages lack explicit head-pose conditioning, and missed faces
remain unmodified. Our approach combines promptable segmentation
with body-derived head localization and guides full-head generation
using depth and projected keypoints. Inpainting starts from the
blurred region, while a synthetic identity reference shared across
each track encourages consistent appearance.

Commercial products, including Celantur, brighter~AI, and
dSPACE UAI Anonymizer, offer cloud-based or on-premise
anonymization~\cite{celantur}. Cloud processing requires transferring
raw personal data to a service provider, while on-premise deployment
keeps processing local.

\textbf{Foundation models for data processing.}
Promptable segmentation (SAM~\cite{kirillov2023sam},
SAM~2~\cite{ravi2025sam2}) and open-vocabulary detection
(OWL-ViT~\cite{minderer2022owlvit}, Grounding DINO~\cite{liu2024groundingdino})
turned many labeling steps into prompting. SAM~3~\cite{carion2026sam3}
segments all instances of a text concept and is the backbone of our
pipeline. Human mesh recovery such as SAM~3D Body~\cite{yang2026sam3dbody}
with the MHR body model~\cite{ferguson2025mhr} recovers a full 3D body
from a single image. In driving, VLMs and foundation models already
speed up labeling, e.g., by inferring vehicle dimensions for 3D
boxes~\cite{chen2026vlmlabeling}.
Anonymization with foundation models and no task-specific training
has been shown with text-to-image generators~\cite{piano2024gem}; for
street-level imagery, CAIAMAR combines open-vocabulary segmentation,
zoomed crops, and multi-agent reasoning~\cite{aufschlager2026caiamar}.
The closest system to ours, the Autoware rosbag2
anonymizer~\cite{autowareanon}, prompts Grounding DINO, verifies boxes
with OpenCLIP, segments them with SAM~2, can require plates to lie
inside vehicles, and distills its output into a YOLO detector; it has
no notion of head orientation and no zoomed search inside vehicles.

\section{Method}
\label{sec:method}

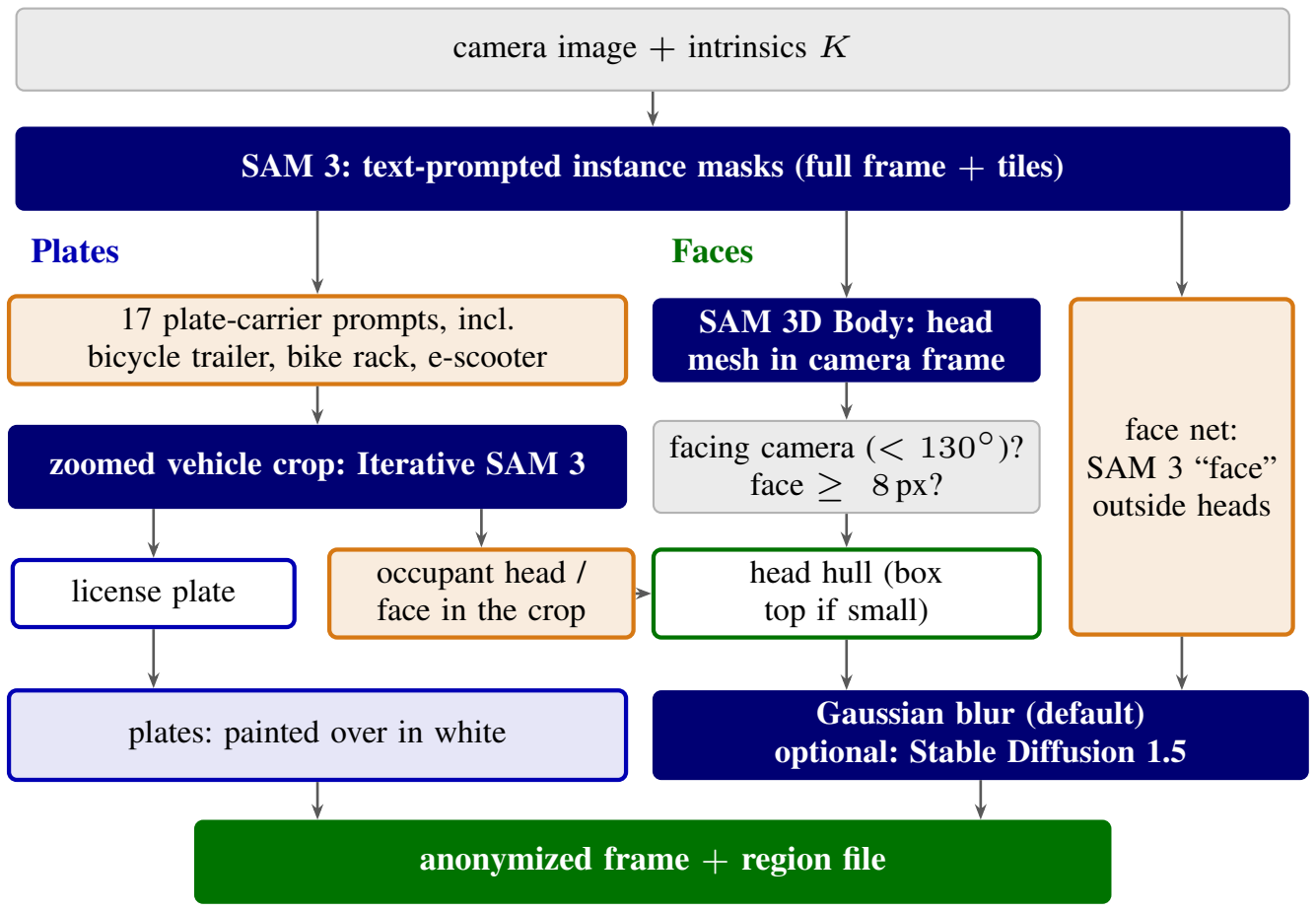
\begin{figure}[t]
\centering
\providecolor{safetynet}{RGB}{214,120,20}
\colorlet{pipenavy}{blue!45!black}
\colorlet{pipegrey}{black!8}
\colorlet{pipeplate}{blue!70!black}
\colorlet{pipeout}{green!45!black}
\begin{tikzpicture}[
  x=1cm, y=1cm,
  font=\fontsize{6.2}{7.2}\selectfont,
  every node/.style={align=center, inner sep=2pt},
  proc/.style={draw=pipenavy, fill=pipenavy, text=white, rounded corners=1.5pt,
    minimum height=0.55cm, font=\fontsize{6.2}{7.2}\selectfont\bfseries},
  ask/.style={draw=black!30, fill=pipegrey, rounded corners=1.5pt,
    minimum height=0.55cm},
  net/.style={draw=safetynet, fill=safetynet!14, line width=0.8pt,
    rounded corners=1.5pt, minimum height=0.55cm},
  hit/.style={draw=#1, fill=white, line width=0.8pt, rounded corners=1.5pt,
    minimum height=0.45cm},
  fin/.style={draw=#1, fill=#1!10, line width=0.8pt, rounded corners=1.5pt,
    minimum height=0.6cm},
  arr/.style={-{Stealth[length=3pt]}, draw=black!65, line width=0.5pt},
  lab/.style={font=\fontsize{5.4}{6}\selectfont\itshape, text=black!65},
]
\node[ask, text width=8.4cm] (in) at (4.35, 0)
  {camera image $+$ intrinsics $K$};
\node[proc, text width=8.4cm] (sam) at (4.35, -0.8)
  {SAM~3: text-prompted instance masks (full frame $+$ tiles)};
\node[font=\fontsize{6.6}{7}\selectfont\bfseries, text=pipeplate, anchor=west]
  at (0.1, -1.35) {Plates};
\node[font=\fontsize{6.6}{7}\selectfont\bfseries, text=pipeout, anchor=west]
  at (4.4, -1.35) {Faces};
\node[net, text width=4.0cm] (carriers) at (2.1, -1.95)
  {17 plate-carrier prompts, incl.\ bicycle trailer, bike rack, e-scooter};
\node[proc, text width=4.0cm] (crop) at (2.1, -2.8)
  {zoomed vehicle crop: Iterative SAM~3};
\node[hit=pipeplate, text width=1.75cm] (plate) at (1.0, -3.65)
  {license plate};
\node[net, text width=1.9cm] (occ) at (3.2, -3.65)
  {occupant head / face in the crop};
\node[proc, text width=2.45cm] (mesh) at (5.65, -1.95)
  {SAM~3D Body: head mesh in camera frame};
\node[ask, text width=2.45cm] (face) at (5.65, -2.8)
  {facing camera ($<130^\circ$)?\\ face $\geq 8$\,px?};
\node[hit=pipeout, text width=2.45cm] (head) at (5.65, -3.65)
  {head hull (box top if small)};
\node[net, text width=1.35cm, minimum height=2.25cm] (facenet)
  at (7.9, -2.8) {face net: SAM~3 ``face'' outside heads};
\node[fin=pipeplate, text width=4.0cm] (white) at (2.1, -4.6)
  {plates: painted over in white};
\node[proc, text width=4.25cm] (diff) at (6.55, -4.6)
  {Gaussian blur (default)\\ optional: Stable Diffusion 1.5};

\node[draw=pipeout, fill=pipeout, text=white, rounded corners=1.5pt,
  minimum height=0.55cm, text width=6.0cm,
  font=\fontsize{6.4}{7.2}\selectfont\bfseries] (out) at (4.35, -5.45)
  {anonymized frame $+$ region file};
\draw[arr] (in) -- (sam);
\draw[arr] (sam.south -| carriers) -- (carriers);
\draw[arr] (sam.south -| mesh) -- (mesh);
\draw[arr] (sam.south -| facenet) -- (facenet);
\draw[arr] (carriers) -- (crop);
\draw[arr] (crop.south -| plate) -- (plate);
\draw[arr] (crop.south -| occ) -- (occ);
\draw[arr] (occ) -- (head);
\draw[arr] (mesh) -- (face);
\draw[arr] (face) -- (head);
\draw[arr] (plate.south) -- (plate.south |- white.north);
\draw[arr] (head.south) -- (head.south |- diff.north);
\draw[arr] (facenet.south) -- (facenet.south |- diff.north);
\draw[arr] (white.south) -- (white.south |- out.north);
\draw[arr] (diff.south) -- (diff.south |- out.north);
\end{tikzpicture}
\caption{Pipeline: SAM~3 segments plate carriers and persons; vehicle
crops are searched for plates and occupants; facing heads come from
the body mesh. Safety nets in orange. Plates are
painted over in white and faces are generated with Stable Diffusion.}
\label{fig:pipeline}
\end{figure}

Our pipeline takes camera images and intrinsics $K$ and uses
pretrained models without task-specific fine-tuning (Fig.~\ref{fig:pipeline}).
Region masks, types, and scores are stored for
inspection and re-rendering without repeating localization.

\textbf{A. Prompted segmentation.} SAM~3 returns instance masks for a
text concept. Because it resizes every input to a $1008\times1008$
square, we run it on the full frame and on two overlapping square tiles
and merge duplicates by mask IoU and containment. The prompts are
\emph{person}, \emph{face}, and seventeen concepts that may carry a plate:
car, van, truck, bus, trailer, tram, motorcycle, moped, scooter,
e-scooter, cargo bike, delivery bike, and the attachments
bicycle trailer, bike trailer, bike rack, caravan, and tractor.

\textbf{B. License plates inside vehicles.} For every vehicle wider than
60\,px, SAM~3 is prompted with \emph{license plate} on a zoomed crop of
the vehicle box, which up-samples small plates to the model's input
resolution. Plates found on the full frame are kept only inside a
vehicle. Each plate mask is dilated by a quarter of its short side, so
the white fill follows the plate's own shape instead of a rectangle.

\textbf{C. Faces through the human mesh.} Each person box is passed to
SAM~3D Body together with $K$, which returns MHR mesh vertices and
keypoints in the camera frame. The head's forward axis is the nose seen
from the midpoint of the ears, orthogonalized to the neck--ear axis;
the facing angle $\alpha$ is its angle to the ray from the head to the
camera ($0^\circ$ frontal, $90^\circ$ profile, $180^\circ$ back of the
head). If $\alpha<130^\circ$, the head vertices of the mesh, i.e.,
those whose strongest skinning weight belongs to a head joint (skull,
jaw, eyes) and not to the neck, are projected with $K$, and their
convex hull, grown by 5\,\%, is blurred: hair to chin, but neither
neck nor shoulders, and a profile blurs the visible half of
the face. Beyond $130^\circ$ only the back of the head is visible, which
shows no facial features and stays unblurred, so labelers keep the
context; this facing test is what keeps face precision high. Persons under 60\,px, failed fits, and fits
whose head lies outside the person box fall back to blurring the top
quarter of the box. By default, faces smaller than $8\times 8$\,px are
left unblurred: with a face width of 16\,cm and an inter-pupillary
distance of about 6.3\,cm, already a $20\times 20$\,px face has an inter-pupillary distance
below 8\,px, and human recognition of unfamiliar faces is at chance
already at 10\,px~\cite{merino2025peepers}. This is a configurable
precision setting, not a guarantee against acquaintances or machine
recognition.
The face size follows from the mesh as $f\,w/z$ with a face width
$w=16$\,cm at head depth $z$, or as 13\,\% of the person height when
no mesh is available. Because this is an estimate, the limit of 8\,px
lies well below these 20\,px as a safety margin.

\begin{table*}[t]
\centering
\caption{Quantitative results on our data and PP4AV. We report privacy
recall R, detected recall D, true positives TP, false positives FP, and
false negatives FN. Bold values are the best in the respective column.}
\label{tab:main}
\resizebox{\textwidth}{!}{\begin{tabular}{l ccccc ccccc ccccc ccccc}
\toprule
 & \multicolumn{5}{c}{faces, our data} & \multicolumn{5}{c}{plates, our data} & \multicolumn{5}{c}{PP4AV faces\textsuperscript{\dag}} & \multicolumn{5}{c}{PP4AV plates} \\
\cmidrule(lr){2-6}\cmidrule(lr){7-11}\cmidrule(lr){12-16}\cmidrule(lr){17-21}
Method & R & D & TP & FP & FN & R & D & TP & FP & FN & R & D & TP & FP & FN & R & D & TP & FP & FN \\
\midrule
deface~\cite{deface} & 96.6 & 97.1 & 337 & 414 & 12 & -- & -- & -- & -- & -- & 15.9 & 84.8 & 571 & 5{,}040 & 3{,}026 & -- & -- & -- & -- & -- \\
EgoBlur~\cite{raina2023egoblur} (0.9) & 42.7 & 44.1 & 149 & \textbf{3} & 200 & 40.7 & 41.1 & 248 & \textbf{5} & 362 & 1.7 & 20.8 & 60 & \textbf{110} & 3{,}537 & 13.0 & 15.9 & 1{,}016 & \textbf{16} & 6{,}771 \\
EgoBlur~\cite{raina2023egoblur} (0.5) & 59.9 & 63.3 & 209 & 148 & 140 & 63.8 & 66.1 & 389 & 36 & 221 & 3.8 & 54.2 & 138 & 1{,}554 & 3{,}459 & 34.1 & 43.7 & 2{,}655 & 324 & 5{,}132 \\
U.\ Anon.~\cite{turkcan2026urbananonymizer} & 93.7 & 94.6 & 327 & 486 & 22 & 73.0 & 73.1 & 445 & 50 & 165 & 82.1 & \textbf{99.1} & 2{,}954 & 46{,}097 & 643 & 87.9 & 88.1 & 6{,}845 & 1{,}028 & 942 \\
AW Anon.~\cite{autowareanon} & 10.6 & 19.2 & 37 & 4 & 312 & 34.9 & 54.9 & 213 & 82 & 397 & 2.7 & 38.9 & 96 & 501 & 3{,}501 & 34.5 & 72.3 & 2{,}688 & 1{,}079 & 5{,}099 \\
\midrule
\ours{} (full) & \textbf{97.1} & \textbf{98.0} & \textbf{339} & 149 & \textbf{10} & \textbf{93.4} & \textbf{97.2} & \textbf{570} & 63 & \textbf{40} & \textbf{97.8} & 98.0 & \textbf{3{,}517} & 6{,}439 & \textbf{80} & \textbf{95.5} & \textbf{97.9} & \textbf{7{,}436} & 7{,}333 & \textbf{351} \\
\ours{} (ign.\ rules) & 97.4 & 98.3 & 340 & 675 & 9 & 97.2 & 97.2 & 593 & 61 & 17 & 98.8 & 99.0 & 3{,}554 & 34{,}638 & 43 & 97.0 & 98.0 & 7{,}556 & 7{,}253 & 231 \\
\bottomrule
\end{tabular}
}\\[2pt]
{\footnotesize\raggedright \textsuperscript{\dag}PP4AV face boxes are larger than the regions face detectors output. Hence, here the face regions of every method are enlarged by $1/0.735$ about their centers; 0.735 is the median ratio of a CenterFace~\cite{xu2019centerface} box to its PP4AV box over 1{,}030 matched PP4AV faces. Both \ours{} rows are enlarged like the baselines.\par}
\end{table*}

\begin{figure*}[t]
\centering
\begin{tikzpicture}
\node[inner sep=0pt] {\includegraphics[width=\textwidth]{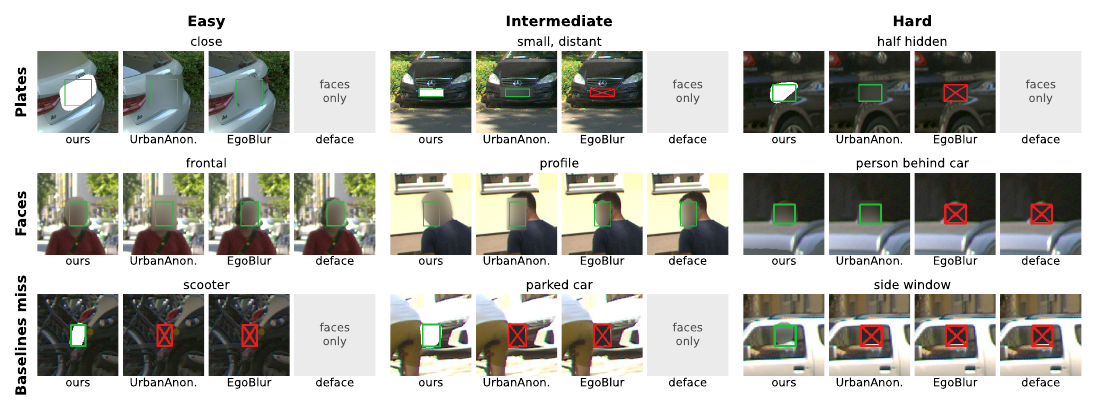}};
\end{tikzpicture}
\caption{Qualitative comparison on plates, faces, and objects the
baselines miss, from easy to hard. Each cell shows every method on the
same crop. Green box: protected; dark,
red-crossed box: missed (original pixels not printed). deface detects
faces only.}
\label{fig:qual}
\end{figure*}

\textbf{D. Safety nets.} SAM~3 \emph{face} detections not covered by
a head region are added; the vehicle crops of step~B are also prompted
with \emph{person} and \emph{face}, so occupants behind the windows are
found and their heads blurred; and the attachment prompts of step~A
catch country-specific plate mountings such as bicycle trailers or
e-scooters.

\textbf{E. Anonymization.} Plates are filled completely white, so no
de-blurring over consecutive frames can recover them. Faces are covered with a Gaussian blur ($\sigma$ = 0.3 times the region's longer
side, feathered only outside the mask). 

\textbf{F. Geometrically guided head generation.}
\label{sec:head-generation}
For each head or occupant region, we run Stable
Diffusion~1.5 inpainting~\cite{rombach2022highresolution}
on a $512\times512$~px crop and replace the whole head
region, dilated by 10\%. Generation starts from the blurred head with denoising
strength $s=0.8$. This retains coarse lighting and color
information while withholding the original head appearance
from the inpainting model.

Where a valid body fit is available, two
ControlNets~\cite{zhang2023controlnet} (v1.1 depth and OpenPose for SD~1.5) guide generation.
The first receives a depth rendering of the MHR template
head at the orientation estimated by SAM~3D Body. Using
the template supplies geometric guidance without reproducing
the person's fitted facial shape. The second receives
the projected face and shoulder keypoints from the body
fit. Regions without a valid fit are generated without
these geometric controls.

\textbf{G. Identity guidance and compositing.}
We link regions of consecutive frames into tracks by box
IoU and assign a synthetic reference face to each track,
generated by Stable Diffusion~1.5 on a grey canvas from a per-track seed and never sampled
from data.
IP-Adapter-FaceID~\cite{ye2023ipadapter} conditions generation
on this reference, encouraging a consistent synthetic
identity as the person's pose and surroundings change.
The generated head is blended into the image using a
Laplacian pyramid and matched to the surrounding image's
blur, noise, and JPEG compression. Only the selected
replacement region is composited into the frame, other
anonymization regions retain their white fill or blur.

\section{Experiments}
\label{sec:exp}

\begin{figure*}[t]
\begin{minipage}[t]{0.635\textwidth}
\centering
\begin{tikzpicture}
\node[inner sep=0pt] {\includegraphics[width=\linewidth]{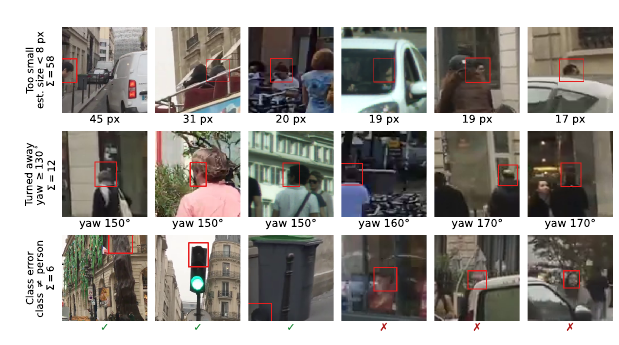}};
\end{tikzpicture}
\caption{In detail on \ppMissRows/\ppMissTotal{} FN of
\ours{} on PP4AV. Size descending, yaw ascending.
Class errors: {\color{green!50!black}\checkmark} label error,
{\color{red!70!black}$\times$} face we miss.}
\label{fig:misses}
\end{minipage}\hfill
\begin{minipage}[t]{0.35\textwidth}
\centering
\begin{tikzpicture}
\node[inner sep=0pt] {\includegraphics[width=\linewidth]{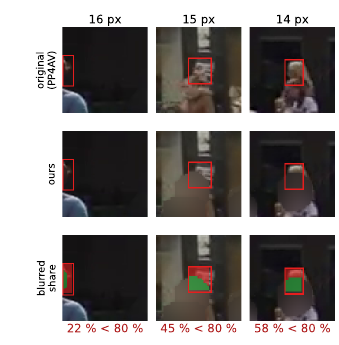}};
\end{tikzpicture}
\caption{3 of the \ppMissFacing/\ppMissTotal{} FN of \ours{} on PP4AV. Bottom: blurred share.}
\label{fig:coverage}
\end{minipage}
\end{figure*}

\subsection{Experimental Setup}
\label{sec:setup}
We run SAM~3~\cite{carion2026sam3} and SAM~3D Body (DINOv3
backbone)~\cite{yang2026sam3dbody} without any fine-tuning on one
RTX~4090, with the settings of Sec.~\ref{sec:method}: score threshold
0.3 for vehicles, persons, and plates and 0.4 for faces, facing limit
$130^\circ$, size limit 8\,px, and $\sigma = 0.3$. All thresholds were
set on our recordings; PP4AV stays held out.
Generation uses the full-head configuration of
Sec.~\ref{sec:method}; heads below 24\,px and other
ineligible regions retain their blur. On an RTX~4090, anonymization takes about 3\,s per frame,
with an additional 4\,s for generation.

\textbf{Datasets.} We use our own recordings for validation and
PP4AV~\cite{trinh2023pp4av} for testing. PP4AV is openly accessible and
holds 3{,}203 frames with 3{,}597 face and 7{,}787 plate boxes; its face
boxes enclose the whole head. Our recordings come from a car on the
street and a micro-mobility agent with a camera on the bicycle lane:
2{,}776 frames of $2200\times1200$\,px, of which we hand-annotated 300
with 349 faces and 610 plates. We box every real face showing eyes,
nose, or mouth, also in profile, behind glass, or partly occluded (no
backs of heads, posters, or screens), and every plate, also unreadable
or cut by the image border.

\textbf{Metrics.} We score the blur that each method applies. An object
is \emph{protected} if at least 80\,\% of its box is blurred; privacy
recall (R) is the share of protected objects. The detected recall (D)
also counts objects that a blur region of their class overlaps with an
IoU of at least 0.3; D measures localization, R the actual privacy. A
blur region on no ground-truth box of its class is a false positive. As
a missed face or plate is irreversible and a false positive only blurs a
few pixels, we rank methods by R, like Street
View~\cite{frome2009streetview} and nuScenes~\cite{caesar2020nuscenes},
and report false positives as its cost. 

We test identity removal with ArcFace
embeddings~\cite{deng2019arcface} and cosine similarity in two attacks. In the first, the attacker
holds the original frames and links each anonymized face crop to a
gallery with one original crop per evaluated face. Rank-1 re-identification is the share of probes whose
closest gallery entry is their own original; unmodified probes reach 100\,\%, lightly blurred ones ($\sigma$ of 4\,\% of the crop side) 49.8 and 57.0\,\%. The gallery often holds other crops of the same person
(for 62\,\% of the faces on our data and 21\,\% on PP4AV, by IoU tracks), so we also report a person-level
rank-1 that counts a match to any crop of the probe's track. The true
acceptance rate (TAR) is the share of probes whose similarity to their own
original exceeds a threshold, set per method so
that 0.1\,\% of all non-matching pairs of the set exceed it
(FAR = 0.1\,\%). In the second attack the gallery holds a later, unmodified
observation: on our data, 118 faces whose person box an IoU
tracker follows are matched against their 118 crops five frames later. Missed
faces remain in both tests; generated faces are
also scored separately.

To measure perception utility, Keypoint
R-CNN~\cite{he2017maskrcnn} predictions on the original
images (score $\geq0.5$) serve as reference annotations.
We report COCO-style average precision
(AP), averaged over
thresholds from 0.50 to 0.95 in steps of 0.05.
Keypoint AP uses object keypoint
similarity (OKS). We report all 17 keypoints
and the five head keypoints separately. Higher AP
indicates better preservation of the original model
predictions, not accuracy against manually annotated
poses. Head-pose preservation is the median
angular difference between facing directions estimated
by SAM~3D Body~\cite{yang2026sam3dbody} on the original
and anonymized images. 

We measure head-crop realism using Kernel Inception
Distance (KID)~\cite{binkowski2018demystifying},
reported as KID$\times1000$; lower values indicate
closer agreement with the original head distribution.

\textbf{Baselines.} We compare against EgoBlur~\cite{raina2023egoblur},
UrbanAnonymizer~\cite{turkcan2026urbananonymizer}, the Autoware
anonymizer~\cite{autowareanon}, and the face-only deface~\cite{deface},
all with their default settings on single frames. EgoBlur, UrbanAnonymizer, and deface enlarge their blurred
boxes by 15, 20, and 30\,\%. EgoBlur's default detection threshold of 0.9
misses many objects, so we also run it at 0.5. For generation, we compare the blur-only output, our full-head
replacement, and our LDFA reproduction~\cite{klemp2023ldfa}.
LDFA uses its own RetinaFace detections and SD~2 inpainting
with no prompt, CFG scale 1, Euler ancestral sampling, 50 steps,
32\,px context padding, and denoising strength 0.75. We use
mirrored pretrained weights and the same JPEG-saving procedure
as our method.

\begin{table}[t]
\centering
\caption{Ablation on our data and PP4AV; arrows against
the full method (green: better, red: worse).}
\label{tab:ablation}
\resizebox{\columnwidth}{!}{\begin{tabular}{l ccccc ccccc}
\toprule
 & \multicolumn{5}{c}{faces} & \multicolumn{5}{c}{plates} \\
\cmidrule(lr){2-6}\cmidrule(lr){7-11}
Variant & R & D & TP & FP & FN & R & D & TP & FP & FN \\
\midrule
\multicolumn{11}{l}{\textbf{our data}} \\
\ours{} (full) & 97.1 & 98.0 & 339 & 149 & 10 & 93.4 & 97.2 & 570 & 63 & 40 \\
\midrule
w/o 8\,px face limit & 97.4\,{\color{green!50!black}$\uparrow$} & 98.3\,{\color{green!50!black}$\uparrow$} & 340\,{\color{green!50!black}$\uparrow$} & 675\,{\color{red!75!black}$\uparrow$} & 9\,{\color{green!50!black}$\downarrow$} & 93.4 & 97.2 & 570 & 63 & 40 \\
w/o person route & 92.3\,{\color{red!75!black}$\downarrow$} & 94.6\,{\color{red!75!black}$\downarrow$} & 322\,{\color{red!75!black}$\downarrow$} & 126\,{\color{green!50!black}$\downarrow$} & 27\,{\color{red!75!black}$\uparrow$} & 93.4 & 97.2 & 570 & 63 & 40 \\
w/o facing test & 97.4\,{\color{green!50!black}$\uparrow$} & 98.3\,{\color{green!50!black}$\uparrow$} & 340\,{\color{green!50!black}$\uparrow$} & 352\,{\color{red!75!black}$\uparrow$} & 9\,{\color{green!50!black}$\downarrow$} & 93.4 & 97.2 & 570 & 63 & 40 \\
facing threshold $90^\circ$ & 94.8\,{\color{red!75!black}$\downarrow$} & 96.3\,{\color{red!75!black}$\downarrow$} & 331\,{\color{red!75!black}$\downarrow$} & 202\,{\color{red!75!black}$\uparrow$} & 18\,{\color{red!75!black}$\uparrow$} & 93.4 & 97.2 & 570 & 63 & 40 \\
w/o tiles & 94.6\,{\color{red!75!black}$\downarrow$} & 95.4\,{\color{red!75!black}$\downarrow$} & 330\,{\color{red!75!black}$\downarrow$} & 137\,{\color{green!50!black}$\downarrow$} & 19\,{\color{red!75!black}$\uparrow$} & 89.3\,{\color{red!75!black}$\downarrow$} & 93.1\,{\color{red!75!black}$\downarrow$} & 545\,{\color{red!75!black}$\downarrow$} & 64\,{\color{red!75!black}$\uparrow$} & 65\,{\color{red!75!black}$\uparrow$} \\
w/o zoomed vehicle crops & 93.4\,{\color{red!75!black}$\downarrow$} & 94.3\,{\color{red!75!black}$\downarrow$} & 326\,{\color{red!75!black}$\downarrow$} & 79\,{\color{green!50!black}$\downarrow$} & 23\,{\color{red!75!black}$\uparrow$} & 81.5\,{\color{red!75!black}$\downarrow$} & 85.1\,{\color{red!75!black}$\downarrow$} & 497\,{\color{red!75!black}$\downarrow$} & 35\,{\color{green!50!black}$\downarrow$} & 113\,{\color{red!75!black}$\uparrow$} \\
\midrule
\multicolumn{11}{l}{\emph{safety nets}} \\
w/o face net & 75.4\,{\color{red!75!black}$\downarrow$} & 76.5\,{\color{red!75!black}$\downarrow$} & 263\,{\color{red!75!black}$\downarrow$} & 118\,{\color{green!50!black}$\downarrow$} & 86\,{\color{red!75!black}$\uparrow$} & 93.4 & 97.2 & 570 & 63 & 40 \\
w/o occupant net & 97.1 & 97.4\,{\color{red!75!black}$\downarrow$} & 339 & 79\,{\color{green!50!black}$\downarrow$} & 10 & 93.4 & 97.2 & 570 & 63 & 40 \\
w/o window blur & 94.3\,{\color{red!75!black}$\downarrow$} & 95.1\,{\color{red!75!black}$\downarrow$} & 329\,{\color{red!75!black}$\downarrow$} & 210\,{\color{red!75!black}$\uparrow$} & 20\,{\color{red!75!black}$\uparrow$} & 93.3\,{\color{red!75!black}$\downarrow$} & 97.0\,{\color{red!75!black}$\downarrow$} & 569\,{\color{red!75!black}$\downarrow$} & 63 & 41\,{\color{red!75!black}$\uparrow$} \\
w/o attachment prompts & 97.1 & 98.0 & 339 & 148\,{\color{green!50!black}$\downarrow$} & 10 & 93.1\,{\color{red!75!black}$\downarrow$} & 96.9\,{\color{red!75!black}$\downarrow$} & 568\,{\color{red!75!black}$\downarrow$} & 59\,{\color{green!50!black}$\downarrow$} & 42\,{\color{red!75!black}$\uparrow$} \\
\midrule[\heavyrulewidth]
\multicolumn{11}{l}{\textbf{PP4AV}} \\
\ours{} (full) & 97.8 & 98.0 & 3{,}517 & 6{,}439 & 80 & 95.5 & 97.9 & 7{,}436 & 7{,}333 & 351 \\
\midrule
w/o 8\,px face limit & 98.8\,{\color{green!50!black}$\uparrow$} & 99.0\,{\color{green!50!black}$\uparrow$} & 3{,}554\,{\color{green!50!black}$\uparrow$} & 34{,}638\,{\color{red!75!black}$\uparrow$} & 43\,{\color{green!50!black}$\downarrow$} & 95.5 & 97.9 & 7{,}436 & 7{,}333 & 351 \\
w/o person route & 91.8\,{\color{red!75!black}$\downarrow$} & 95.5\,{\color{red!75!black}$\downarrow$} & 3{,}303\,{\color{red!75!black}$\downarrow$} & 4{,}343\,{\color{green!50!black}$\downarrow$} & 294\,{\color{red!75!black}$\uparrow$} & 95.5 & 97.9 & 7{,}436 & 7{,}333 & 351 \\
w/o facing test & 98.1\,{\color{green!50!black}$\uparrow$} & 98.3\,{\color{green!50!black}$\uparrow$} & 3{,}530\,{\color{green!50!black}$\uparrow$} & 9{,}332\,{\color{red!75!black}$\uparrow$} & 67\,{\color{green!50!black}$\downarrow$} & 95.5 & 97.9 & 7{,}436 & 7{,}333 & 351 \\
facing threshold $90^\circ$ & 96.4\,{\color{red!75!black}$\downarrow$} & 97.3\,{\color{red!75!black}$\downarrow$} & 3{,}467\,{\color{red!75!black}$\downarrow$} & 6{,}028\,{\color{green!50!black}$\downarrow$} & 130\,{\color{red!75!black}$\uparrow$} & 95.5 & 97.9 & 7{,}436 & 7{,}333 & 351 \\
w/o tiles & 95.9\,{\color{red!75!black}$\downarrow$} & 96.2\,{\color{red!75!black}$\downarrow$} & 3{,}450\,{\color{red!75!black}$\downarrow$} & 5{,}081\,{\color{green!50!black}$\downarrow$} & 147\,{\color{red!75!black}$\uparrow$} & 93.0\,{\color{red!75!black}$\downarrow$} & 95.4\,{\color{red!75!black}$\downarrow$} & 7{,}244\,{\color{red!75!black}$\downarrow$} & 5{,}674\,{\color{green!50!black}$\downarrow$} & 543\,{\color{red!75!black}$\uparrow$} \\
w/o zoomed vehicle crops & 97.7\,{\color{red!75!black}$\downarrow$} & 97.9\,{\color{red!75!black}$\downarrow$} & 3{,}514\,{\color{red!75!black}$\downarrow$} & 5{,}838\,{\color{green!50!black}$\downarrow$} & 83\,{\color{red!75!black}$\uparrow$} & 95.5 & 97.1\,{\color{red!75!black}$\downarrow$} & 7{,}435\,{\color{red!75!black}$\downarrow$} & 4{,}886\,{\color{green!50!black}$\downarrow$} & 352\,{\color{red!75!black}$\uparrow$} \\
\midrule
\multicolumn{11}{l}{\emph{safety nets}} \\
w/o face net & 83.3\,{\color{red!75!black}$\downarrow$} & 83.4\,{\color{red!75!black}$\downarrow$} & 2{,}996\,{\color{red!75!black}$\downarrow$} & 3{,}711\,{\color{green!50!black}$\downarrow$} & 601\,{\color{red!75!black}$\uparrow$} & 95.5 & 97.9 & 7{,}436 & 7{,}333 & 351 \\
w/o occupant net & 97.7\,{\color{red!75!black}$\downarrow$} & 97.9\,{\color{red!75!black}$\downarrow$} & 3{,}516\,{\color{red!75!black}$\downarrow$} & 5{,}838\,{\color{green!50!black}$\downarrow$} & 81\,{\color{red!75!black}$\uparrow$} & 95.5 & 97.9 & 7{,}436 & 7{,}333 & 351 \\
w/o window blur & 97.8 & 98.0 & 3{,}517 & 6{,}439 & 80 & 95.4\,{\color{red!75!black}$\downarrow$} & 97.9 & 7{,}432\,{\color{red!75!black}$\downarrow$} & 7{,}333 & 355\,{\color{red!75!black}$\uparrow$} \\
w/o attachment prompts & 97.8 & 98.0 & 3{,}517 & 6{,}425\,{\color{green!50!black}$\downarrow$} & 80 & 95.5 & 97.9 & 7{,}436 & 7{,}315\,{\color{green!50!black}$\downarrow$} & 351 \\
\bottomrule
\end{tabular}
}
\end{table}
\textbf{Detection Ablation.} Table~\ref{tab:ablation} removes one
component at a time; numbers are given as our data / PP4AV, and the
trends agree unless noted. Dropping the 8\,px face limit saves a few
faces but multiplies the false positives by about five. The person route, which integrates SAM~3D Body,
improves every face column except FP: without it, R drops, while it costs only 23 / 2{,}096 false positives. Its facing
threshold of $130^\circ$ is the right counterpart: without the test,
face false positives rise, while
$90^\circ$ is too low and leaves more identifiable faces
unblurred. SAM~3D Body also provides the head pose and keypoints for
the generative replacement of Sec.~\ref{sec:gen-results}. The two
overlapping image tiles are passed to SAM~3 at full resolution, avoiding
its downscaling, adding both faces and plates. The zoomed
vehicle crops are the main enabler for plates on our data but on PP4AV they add little benefit at a high cost due to its lower resolution. Among the safety nets of
Fig.~\ref{fig:pipeline}, the dedicated face prompt catches heavily
occluded people (10 instead of 86 / 80 instead of 601 missed faces).
The occupant net registers vehicle occupants in the zoomed car
segmentation; together with the window blur, which protects 10
occupants behind reflections on our data, it covers people inside
vehicles. Lastly, the attachment prompts encode country-specific
traffic and plate rules, a quick, training-free adaptation to the use
case. They improve nearly every plate column on our data but transfer
poorly to other countries and bring no gain on PP4AV.

\begin{table*}[!t]
\centering
\caption{Generative replacement compared with blur and our LDFA
reproduction on all annotated faces of at least
20\,px: 225 of the 349 in our 300 frames and 1{,}136 of the 3{,}597 on PP4AV (648 images); each is one probe and one gallery entry.
ArcFace embeds unaligned square crops of $1.3\times$ the annotated face box
at $112\times112$\,px, identically for
all methods and the gallery of originals, so no probe is dropped. Blur and \ours{}-Gen modify the head
regions of the Table~\ref{tab:main} runs (without the enlargement on
PP4AV); LDFA modifies its RetinaFace
detections. Keypoint AP: 227 and 2{,}150 persons with an anonymized head; facing: 268 and 2{,}163 heads; KID ($\times1000$): 340 and 2{,}407 head crops.}
\label{tab:generation}
\small
\setlength{\tabcolsep}{6pt}
\begin{tabular}{llrrrrrr}
\hline
 & & \multicolumn{2}{c}{Re-identification $\downarrow$}
 & KID $\downarrow$
 & \multicolumn{2}{c}{Keypoint AP $\uparrow$}
 & Facing $\downarrow$ \\
Data & Method & Rank-1 & TAR & & All & Head & $(^\circ)$ \\
\hline
Our data

 & LDFA (reprod.)   & 64.9 & 24.9 & 0.1   & 94.3 & 90.4 & 1.1 \\
 & \ours{} (blur)      & 4.0  & 4.0  & 109.7 & 54.1 & 22.1 & 5.4 \\
 & \ours{}-Gen     & 6.2  & 4.4  & 14.4  & 79.6 & 69.7 & 3.3 \\
 & \quad w/o ControlNets & 4.4 & 3.1 & 13.9 & 57.5 & 25.2 & 13.9 \\
\hline
PP4AV
 & LDFA (reprod.)   & 78.5 & 49.2 & 0.1   & 89.7 & 83.4 & 1.5 \\
  & \ours{} (blur)      & 3.3  & 4.0  & 97.4  & 39.9 & 11.7 & 9.9 \\
 & \ours{}-Gen     & 6.4  & 4.4  & 9.9   & 68.3 & 54.2 & 4.4 \\
\hline
\end{tabular}
\end{table*}

\subsection{Evaluation}
\label{sec:eval}

We compare \ours{} with the four baselines quantitatively in
Table~\ref{tab:main} and qualitatively in Fig.~\ref{fig:qual}.

\textbf{Faces.} On our data, \ours{} protects 97.1\,\% of the faces, on
par with the strongest baseline, deface (96.6\,\%). On PP4AV,
\ours{} protects 97.8\,\%; only UrbanAnonymizer comes close, 15.7 points
lower in R but 1.1 points higher in D (99.1\,\%): it finds the faces,
but despite its 20\,\% padding it covers 17\,\% of them by less than
80\,\%. Of our 80 misses on PP4AV (Figs.~\ref{fig:misses}
and~\ref{fig:coverage}), 58 have an estimated face size below 8\,px,
12 are turned away, 6 are class errors, and 4 are blurred by less than
80\,\%. We consider faces below 8\,px too small for identification;
without this limit, \ours{} (ign.\ rules) misses 37 faces fewer but
adds 28{,}199 false positives. The turned-away faces in
Fig.~\ref{fig:misses} have a yaw of 150 to $170^\circ$, clearly beyond
the $130^\circ$ limit. Of the six class errors, three are label errors
and three are faces we miss. We keep the head region tight for the face generation of Sec.~\ref{sec:head-generation}. Overall, \ours{} reaches its recall with far fewer false positives
than deface and UrbanAnonymizer on our data (149 vs.\ 414 and 486) and
than UrbanAnonymizer on PP4AV (6{,}439 vs.\ 46{,}097).

\textbf{Plates.} On our data, \ours{} protects 93.4\,\% of the plates,
20.4 points more than the strongest baseline, UrbanAnonymizer
(73.0\,\%), which misses mostly small plates: below 16\,px it protects
27.3\,\% against our 95.1\,\%. On PP4AV, \ours{} protects 95.5\,\%
against 87.9\,\% (below 16\,px: 93.8 vs.\ 74.4\,\%). More than half of
our 40 misses on our data are large, steep plates: we fill a plate in
its own shape, which leaves the corners of its axis-aligned box
uncovered, so plates above 64\,px reach only 76.1\,\%. Filling the whole
box instead (ign.\ rules) raises R to 97.2\,\% on our data and
97.0\,\% on PP4AV without additional false positives; R then equals D,
so the remaining 17 misses are plates that are not found at all. Our
plate false positives are on par with UrbanAnonymizer on our data
(63 vs.\ 50) but higher on PP4AV (7{,}333 vs.\ 1{,}028).

\begin{figure}[!t]
\centering
\includegraphics[width=\columnwidth]{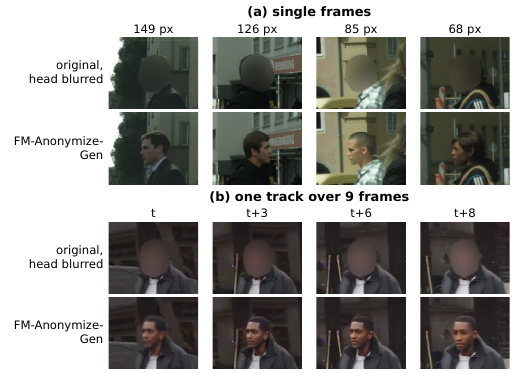}
\caption{Generative head replacement on PP4AV.
(a) Original and generated heads at different sizes and
orientations. (b) Four observations from a nine-frame track,
using a shared synthetic identity reference. The examples
illustrate changes in facial appearance and hair while
approximately retaining head direction and scene lighting.}
\label{fig:generation}
\end{figure}

\subsection{Generative replacement}
\label{sec:gen-results}

\textbf{Perception utility.}
Generated heads improve perception utility over
blur (Table~\ref{tab:generation}). Full-body keypoint AP rises
from 54.1 to 79.6 on our recordings and from 39.9 to 68.3 on
PP4AV. Head-keypoint AP increases from 22.1 to 69.7 and from
11.7 to 54.2, respectively. Estimated head direction stays
closer to the original, and KID decreases on both datasets.

\textbf{Identity removal and comparison with LDFA.}
Our LDFA reproduction leaves
substantial identity information recoverable: rank-1
re-identification reaches 64.9\,\% on our recordings and
78.5\,\% on PP4AV. Its detector misses 38 of 225 and 143 of 1{,}136 faces, which stay original and are re-identified (38 and 142), but among the faces it processes the rate is still 57.8\,\% and 75.5\,\%. Started from noise (strength 1.0), it still reaches 88.4\,\% on the 155-face PP4AV subset below (91.0\,\% at 0.75). Its higher keypoint AP and lower pose
deviation and KID therefore accompany weak identity removal
under our evaluation, which compares two systems with different detectors and regions, not two generators. Our replacement reduces rank-1
re-identification to 6.2\,\% and 6.4\,\%, respectively, while
improving keypoint AP over blur.

Rank-1 remains higher than for blur (4.0\,\% and 3.3\,\%; 14 vs.\ 9 of 225 and 73 vs.\ 37 of 1{,}136 hits). At the person level it is 8.4\,\% and 6.8\,\%, against 4.4\,\% and 3.3\,\% for blur and 72.9\,\% and 80.6\,\% for LDFA.
Of the evaluated faces, 174 (77.3\,\%) and 1{,}010 (88.9\,\%) are generated, 48 and 103 keep their blur, and 3 and 23 are missed. On the generated faces alone, rank-1 is 5.7 vs.\ 2.9\,\% for blur on our data (10 vs.\ 5 hits) and 4.2 vs.\ 0.8\,\% on PP4AV (42 vs.\ 8); the faces that keep their blur add 1 and 11 hits (blur: 1 and 9), the missed ones 3 and 20.
As our inpainting has no
direct access to the original facial texture, a correct
gallery match does not by itself demonstrate reconstruction, but it still indicates recoverable identity cues, also from residual appearance
and context. TAR at 0.1\,\% FAR changes from 4.0\,\% to 4.4\,\% on both datasets. Against the later observation (118 probes of 20 to 101\,px, chance 0.8\,\%), rank-1
is 6.8\,\% (8) for generated heads and 2.5\,\% (3) after blurring. Unmodified probes and our LDFA reproduction both reach 18.6\,\% (22), which bounds this attack: the faces are small and unaligned, and the later box is propagated with the person box.

\textbf{Qualitative results and temporal consistency.}
The shared synthetic reference encourages continuity across
the track in Fig.~\ref{fig:generation}(b), but independent
frame generation permits changes in identity and fine detail.
On three tracks per dataset (85 and 20 frame pairs), the median ArcFace cosine similarity of consecutive faces is 0.798 after generation against 0.813 before on our data, and 0.843 against 0.928 on PP4AV; this illustrates appearance consistency, not privacy.

\textbf{Guidance Ablation.}
Disabling both geometric ControlNets (Table~\ref{tab:generation})
keeps KID low, but the facing error exceeds that of blur, and
head-keypoint AP and rank-1 re-identification fall to near blur. The same holds on a 125-image PP4AV subset: facing error $16.7^\circ$ (blur $9.7^\circ$, with ControlNets
$4.2^\circ$), head-keypoint AP 11.2\,\% (11.6, 53.1), and rank-1
6.5\,\% (7.1, 14.8 $[8.7, 22.1]$; 6.4\,\% on the full set with its 1{,}136-face gallery). Geometric guidance thus preserves
pose and keypoint utility.

\subsection{Limitations and Rules that Need Attention}
What has to be anonymized, and where plates are mounted, depends on the
country. Our prompts follow German traffic: mopeds and e-scooters carry
small insurance plates, bike racks and trailers repeat the plate of the
towing car, and caravans and tractors carry their own. Other countries
need other rules, e.g., some US states have no front plates, and taxi
or fleet numbers, company names, or phone numbers on vans can identify
a person as well. With promptable segmentation, each such rule is one
more text prompt instead of a retrained detector: the single prompt
\emph{text} on the vehicle crops already finds company names on vans
and brand names on bicycles and e-scooters. Faces in mirrors are found
only when large enough, and learned deblurring and context-based
re-identification~\cite{mcpherson2016defeating,oh2016faceless} are
untested. SAM~3 and SAM~3D Body use the SAM License, DINOv3 its own,
both with gated weights.

\FloatBarrier
\section{Conclusion}
\ours{} anonymizes driving data without training by instructing
promptable segmentation and a human mesh model. On PP4AV it protects
97.8\,\% of the faces and 95.5\,\% of the plates, on our recordings
97.1\,\% and 93.4\,\%, ahead of all open-source baselines on plates and
on par or ahead on faces, with fewer face false positives than the
closest baselines. The head geometry of SAM~3D Body also grounds a
generative replacement that raises full-body keypoint AP over blur from
54.1 to 79.6\,\% (PP4AV: 39.9 to 68.3\,\%) while re-identification stays near blur and far below the generative competitor (rank-1 6.2\,\% vs.\ 4.0\,\% and 64.9\,\% for LDFA).
Country-specific plate and text rules can be added as prompts; future
work will add temporal tracking and distill the pipeline into a fast
detector.

% ...

\bibliographystyle{IEEEtran}
\bibliography{refs}

@IEEEtranBSTCTL{IEEEexample:BSTcontrol,
  CTLuse_forced_etal       = "yes",
  CTLmax_names_forced_etal = "5",
  CTLnames_show_etal       = "1"
}

@misc{xu2019centerface,
  author={Xu, Yuanyuan and Yan, Wan and Yang, Genke and Luo, Jiliang and Li, Tao and He, Jianan},
  title        = {{CenterFace}: Joint Face Detection and Alignment Using Face as Point},
  howpublished = {Scientific Programming},
  year         = {2020}
}

@misc{deface,
  author       = {Drawitsch, Martin and others},
  title        = {deface: Video anonymization by face detection},
  howpublished = {\url{https://github.com/ORB-HD/deface}},
  year         = {2023}
}

@misc{raina2023egoblur,
  author       = {Raina, Nikhil and Somasundaram, Guruprasad and Zheng, Kang and Miglani, Sagar and Saarinen, Steve and Meissner, Jeff and Schwesinger, Mark and Pesqueira, Luis and Prasad, Ishita and Miller, Edward and Gupta, Prince and Yan, Mingfei and Newcombe, Richard and Ren, Carl and Parkhi, Omkar M.},
  title        = {{EgoBlur}: Responsible Innovation in {Aria}},
  howpublished = {arXiv:2308.13093},
  year         = {2023}
}

@misc{uaianonymizer,
  author       = {{understand.ai}},
  title        = {anonymizer: An anonymizer to obfuscate faces and license plates},
  howpublished = {\url{https://github.com/understand-ai/anonymizer}},
  year         = {2018}
}

@inproceedings{hukkelas2019deepprivacy,
  author    = {Hukkel{\aa}s, H{\aa}kon and Mester, Rudolf and Lindseth, Frank},
  title     = {{DeepPrivacy}: A Generative Adversarial Network for Face Anonymization},
  booktitle = {\isvc},
  year      = {2019}
}

@inproceedings{hukkelas2023deepprivacy2,
  author    = {Hukkel{\aa}s, H{\aa}kon and Lindseth, Frank},
  title     = {{DeepPrivacy2}: Towards Realistic Full-Body Anonymization},
  booktitle = {\wacv},
  year      = {2023}
}

@inproceedings{hukkelas2023doesanon,
  author    = {Hukkel{\aa}s, H{\aa}kon and Lindseth, Frank},
  title     = {Does Image Anonymization Impact Computer Vision Training?},
  booktitle = {\cvpr{} Workshops (Workshop on Autonomous Driving)},
  year      = {2023}
}

@inproceedings{frome2009streetview,
  author    = {Frome, Andrea and Cheung, German and Abdulkader, Ahmad and Zennaro, Marco and Wu, Bo and Bissacco, Alessandro and Adam, Hartwig and Neven, Hartmut and Vincent, Luc},
  title     = {Large-scale Privacy Protection in {Google Street View}},
  booktitle = {\iccv},
  year      = {2009}
}

@inproceedings{klemp2023ldfa,
  author    = {Klemp, Marvin and R{\"o}sch, Kevin and Wagner, Royden and Quehl, Jannik and Lauer, Martin},
  title     = {{LDFA}: Latent Diffusion Face Anonymization for Self-driving Applications},
  booktitle = {\cvpr{} Workshops},
  year      = {2023}
}

@article{wang2026diffanon,
  author  = {Wang, Tong and Xie, Hui and Gao, Feng and Meng, Zian and Zhang, Pengcheng and Duan, Guohao},
  title   = {Diffusion-Based Anonymization and Foundation Model-Powered Semi-Automatic Image Annotation for Privacy-Protective Intelligent Connected Vehicle Traffic Data},
  journal = {World Electric Vehicle Journal},
  year    = {2026}
}

@inproceedings{trinh2023pp4av,
  author    = {Trinh, Linh and Pham, Phuong and Trinh, Hoang and Bach, Nguyen and Nguyen, Dung and Nguyen, Giang and Nguyen, Huy},
  title     = {{PP4AV}: A Benchmarking Dataset for Privacy-Preserving Autonomous Driving},
  booktitle = {\wacv},
  year      = {2023}
}

@inproceedings{kirillov2023sam,
  author    = {Kirillov, Alexander and Mintun, Eric and Ravi, Nikhila and Mao, Hanzi and Rolland, Chloe and Gustafson, Laura and Xiao, Tete and Whitehead, Spencer and Berg, Alexander C. and Lo, Wan-Yen and Doll{\'a}r, Piotr and Girshick, Ross},
  title     = {Segment Anything},
  booktitle = {\iccv},
  year      = {2023}
}

@inproceedings{ravi2025sam2,
  author    = {Ravi, Nikhila and Gabeur, Valentin and Hu, Yuan-Ting and Hu, Ronghang and Ryali, Chaitanya and Ma, Tengyu and Khedr, Haitham and R{\"a}dle, Roman and Rolland, Chloe and Gustafson, Laura and Mintun, Eric and Pan, Junting and Alwala, Kalyan Vasudev and Carion, Nicolas and Wu, Chao-Yuan and Girshick, Ross and Doll{\'a}r, Piotr and Feichtenhofer, Christoph},
  title     = {{SAM 2}: Segment Anything in Images and Videos},
  booktitle = {\iclr},
  year      = {2025}
}

@inproceedings{carion2026sam3,
  author    = {Carion, Nicolas and Gustafson, Laura and Hu, Yuan-Ting and Debnath, Shoubhik and Hu, Ronghang and Suris, Didac and Ryali, Chaitanya and Alwala, Kalyan Vasudev and Khedr, Haitham and Huang, Andrew and Lei, Jie and Ma, Tengyu and Guo, Baishan and Kalla, Arpit and Marks, Markus and Greer, Joseph and Wang, Meng and Sun, Peize and R{\"a}dle, Roman and Afouras, Triantafyllos and Mavroudi, Effrosyni and Xu, Katherine and Wu, Tsung-Han and Zhou, Yu and Momeni, Liliane and Hazra, Rishi and Ding, Shuangrui and Vaze, Sagar and Porcher, Francois and Li, Feng and Li, Siyuan and Kamath, Aishwarya and Cheng, Ho Kei and Doll{\'a}r, Piotr and Ravi, Nikhila and Saenko, Kate and Zhang, Pengchuan and Feichtenhofer, Christoph},
  title     = {{SAM 3}: Segment Anything with Concepts},
  booktitle = {\iclr},
  year      = {2026}
}

@inproceedings{liu2024groundingdino,
  author    = {Liu, Shilong and Zeng, Zhaoyang and Ren, Tianhe and Li, Feng and Zhang, Hao and Yang, Jie and Jiang, Qing and Li, Chunyuan and Yang, Jianwei and Su, Hang and Zhu, Jun and Zhang, Lei},
  title     = {Grounding {DINO}: Marrying {DINO} with Grounded Pre-Training for Open-Set Object Detection},
  booktitle = {\eccv},
  year      = {2024}
}

@inproceedings{minderer2022owlvit,
  author    = {Minderer, Matthias and Gritsenko, Alexey and Stone, Austin and Neumann, Maxim and Weissenborn, Dirk and Dosovitskiy, Alexey and Mahendran, Aravindh and Arnab, Anurag and Dehghani, Mostafa and Shen, Zhuoran and Wang, Xiao and Zhai, Xiaohua and Kipf, Thomas and Houlsby, Neil},
  title     = {Simple Open-Vocabulary Object Detection with Vision Transformers},
  booktitle = {\eccv},
  year      = {2022}
}

@inproceedings{yang2026sam3dbody,
  author    = {Yang, Xitong and Kukreja, Devansh and Pinkus, Don and Sagar, Anushka and Fan, Taosha and Park, Jinhyung and Shin, Soyong and Cao, Jinkun and Liu, Jiawei and Ugrinovic, Nicolas and Feiszli, Matt and Malik, Jitendra and Doll{\'a}r, Piotr and Kitani, Kris},
  title     = {{SAM 3D Body}: Robust Full-Body Human Mesh Recovery},
  booktitle = {\cvpr},
  year      = {2026}
}

@misc{ferguson2025mhr,
  author       = {Ferguson, Aaron and Osman, Ahmed A. A. and Bescos, Berta and Stoll, Carsten and Twigg, Chris and Lassner, Christoph and Otte, David and Vignola, Eric and Prada, Fabian and Bogo, Federica and Santesteban, Igor and Romero, Javier and Zarate, Jenna and Lee, Jeongseok and Park, Jinhyung and Yang, Jinlong and Doublestein, John and Venkateshan, Kishore and Kitani, Kris and Kavan, Ladislav and Farra, Marco Dal and Hu, Matthew and Cioffi, Matthew and Fabris, Michael and Ranieri, Michael and Modarres, Mohammad and Kadlecek, Petr and Khirodkar, Rawal and Abdrashitov, Rinat and Pr{\'e}vost, Romain and Rajbhandari, Roman and Mallet, Ronald and Pearsall, Russell and Kao, Sandy and Kumar, Sanjeev and Parrish, Scott and Yu, Shoou-I and Saito, Shunsuke and Shiratori, Takaaki and Wang, Te-Li and Tung, Tony and Xu, Yichen and Dong, Yuan and Chen, Yuhua and Xu, Yuanlu and Ye, Yuting and Jiang, Zhongshi},
  title        = {{MHR}: Momentum Human Rig},
  howpublished = {arXiv:2511.15586},
  year         = {2025}
}

@inproceedings{chen2026vlmlabeling,
  author    = {Chen, Steven and Khaitan, Shivesh and Djuric, Nemanja},
  title     = {Improving {3D} Labeling in Self-Driving by Inferring Vehicle Information using Vision Language Models},
  booktitle = {\iv},
  year      = {2026}
}

@inproceedings{caesar2020nuscenes,
  author    = {Caesar, Holger and Bankiti, Varun and Lang, Alex H. and Vora, Sourabh and Liong, Venice Erin and Xu, Qiang and Krishnan, Anush and Pan, Yu and Baldan, Giancarlo and Beijbom, Oscar},
  title     = {{nuScenes}: A Multimodal Dataset for Autonomous Driving},
  booktitle = {\cvpr},
  year      = {2020}
}

@inproceedings{wilson2021argoverse2,
  author    = {Wilson, Benjamin and Qi, William and Agarwal, Tanmay and Lambert, John and Singh, Jagjeet and Khandelwal, Siddhesh and Pan, Bowen and Kumar, Ratnesh and Hartnett, Andrew and Pontes, Jhony Kaesemodel and Ramanan, Deva and Carr, Peter and Hays, James},
  title     = {Argoverse 2: Next Generation Datasets for Self-Driving Perception and Forecasting},
  booktitle = {\nips},
  year      = {2021}
}

@misc{geyer2020a2d2,
  author       = {Geyer, Jakob and Kassahun, Yohannes and Mahmudi, Mentar and Ricou, Xavier and Durgesh, Rupesh and Chung, Andrew S. and Hauswald, Lorenz and Pham, Viet Hoang and M{\"u}hlegg, Maximilian and Dorn, Sebastian and Fernandez, Tiffany and J{\"a}nicke, Martin and Mirashi, Sudesh and Savani, Chiragkumar and Sturm, Martin and Vorobiov, Oleksandr and Oelker, Martin and Garreis, Sebastian and Schuberth, Peter},
  title        = {{A2D2}: {Audi} Autonomous Driving Dataset},
  howpublished = {arXiv:2004.06320},
  year         = {2020}
}

@misc{gdpr2016,
  author       = {{European Parliament and Council of the European Union}},
  title        = {Regulation ({EU}) 2016/679 ({General Data Protection Regulation})},
  howpublished = {Official Journal of the European Union},
  year         = {2016}
}

@misc{aiact2024,
  author       = {{European Parliament and Council of the European Union}},
  title        = {Regulation ({EU}) 2024/1689 laying down harmonised rules on artificial intelligence ({Artificial Intelligence Act})},
  howpublished = {Official Journal of the European Union},
  year         = {2024}
}

@misc{edpb2020video,
  author = {{European Data Protection Board}},
  title  = {Guidelines 3/2019 on processing of personal data through video devices, Version 2.0},
  year   = {2020}
}

@misc{celantur,
  author       = {{Celantur GmbH}},
  title        = {Celantur Cloud {API} and Container: Automated Image Anonymization},
  howpublished = {\url{https://www.celantur.com/api/}},
  year         = {2026}
}

@misc{autowareanon,
  author       = {{Autoware Foundation}},
  title        = {autoware\_rosbag2\_anonymizer},
  howpublished = {\url{https://github.com/autowarefoundation/autoware_rosbag2_anonymizer}},
  year         = {2024}
}

@inproceedings{merino2025peepers,
  author    = {Merino, Xavier and Pangelinan, Gabriella and Langborgh, Samuel and King, Michael C. and Bowyer, Kevin W.},
  title     = {Peepers \& Pixels: Human Recognition Accuracy on Low Resolution Faces},
  booktitle = {\fg},
  year      = {2025}
}

@misc{turkcan2026urbananonymizer,
  author       = {Turkcan, Mehmet Kerem},
  title        = {{UrbanAnonymizer}: Face and License Plate Anonymization for Street-Level Images and Video},
  howpublished = {\url{https://huggingface.co/mehmetkeremturkcan/UrbanAnonymizer}},
  year         = {2026}
}

@misc{mcpherson2016defeating,
  author       = {McPherson, Richard and Shokri, Reza and Shmatikov, Vitaly},
  title        = {Defeating Image Obfuscation with Deep Learning},
  howpublished = {arXiv:1609.00408},
  year         = {2016}
}

@inproceedings{oh2016faceless,
  author    = {Oh, Seong Joon and Benenson, Rodrigo and Fritz, Mario and Schiele, Bernt},
  title     = {Faceless Person Recognition: Privacy Implications in Social Media},
  booktitle = {\eccv},
  year      = {2016}
}

@inproceedings{piano2024gem,
  author    = {Piano, Luca and Basci, Pietro and Lamberti, Fabrizio and Morra, Lia},
  title     = {Harnessing Foundation Models for Image Anonymization},
  booktitle = {\gem},
  year      = {2024}
}

@inproceedings{aufschlager2026caiamar,
  author    = {Aufschl{\"a}ger, Robert and Folz, Jakob and Savaliya, Gautam and Vidanalage, Manjitha D. and Heigl, Michael and Schramm, Martin},
  title     = {Towards Context-Aware Image Anonymization with Multi-Agent Reasoning},
  booktitle = {\cvpr{} Workshops (GRAIL-V)},
  year      = {2026}
}

@inproceedings{zhang2023controlnet,
  author    = {Zhang, Lvmin and Rao, Anyi and Agrawala, Maneesh},
  title     = {Adding Conditional Control to Text-to-Image Diffusion Models},
  booktitle = {\iccv},
  year      = {2023}
}

@misc{ye2023ipadapter,
  author       = {Ye, Hu and Zhang, Jun and Liu, Sibo and Han, Xiao and Yang, Wei},
  title        = {{IP-Adapter}: Text Compatible Image Prompt Adapter for Text-to-Image Diffusion Models},
  howpublished = {arXiv:2308.06721},
  year         = {2023}
}

@inproceedings{he2017maskrcnn,
  author    = {He, Kaiming and Gkioxari, Georgia and Doll{\'a}r, Piotr and Girshick, Ross},
  title     = {Mask {R-CNN}},
  booktitle = {\iccv},
  year      = {2017}
}

@inproceedings{rombach2022highresolution,
  author = {Rombach, Robin and Blattmann, Andreas and Lorenz, Dominik and Esser, Patrick and Ommer, Bj{\"o}rn},
  title = {High-Resolution Image Synthesis with Latent Diffusion Models},
  booktitle = {\cvpr},
  year = {2022},
}

@inproceedings{deng2019arcface,
  author = {Deng, Jiankang and Guo, Jia and Xue, Niannan and Zafeiriou, Stefanos},
  title = {{ArcFace}: Additive Angular Margin Loss for Deep Face Recognition},
  booktitle = {\cvpr},
  year = {2019},
}

@inproceedings{binkowski2018demystifying,
  author = {Bi{\'n}kowski, Miko{\l}aj and Sutherland, Danica J. and Arbel, Michael and Gretton, Arthur},
  title = {Demystifying {MMD GANs}},
  booktitle = {\iclr},
  year = {2018}
}
\end{document}